# Persian Causality Corpus (PerCause) and the Causality Detection Benchmark

Zeinab Rahimi, Mehrnoush ShamsFard*[1]

*NLP Research Lab., Shahid Beheshti University, Tehran, Iran*
{z_rahimi, m-shams}@sbu.ac.ir

*ABSTRACT. Recognizing causal elements and causal relations in text is one of the challenging issues in natural language processing; specifically, in low resource languages such as Persian. In this research we prepare a causality human annotated corpus for the Persian language which consists of 4446 sentences and 5128 causal relations and three labels of cause, effect and causal mark -if possible- are specified for each relation. We have used this corpus to train a system for detecting causal elements' boundaries.*

*Also, we present a causality detection benchmark for three machine learning methods and two deep learning systems based on this corpus. Performance evaluations indicate that our best total result is obtained through CRF classifier which has F-measure of 0.76 and the best accuracy obtained through BiLSTM-CRF deep learning method with Accuracy equal to %91.4.*

Keywords: PerCause, Causality annotated corpus, causality detection, deep learning, CRF

## 1. Introduction

Causal relation extraction is a challenging natural language processing (NLP) open problem that mainly involves semantic analysis and is used in many NLP tasks such as textual entailment recognition, question answering, event prediction and narrative extraction. Causation indicates that one event, state or action is the result of the occurrence of the other state, event or action. Consider the following examples:

(1)   John is poisoned because he ate a poisonous apple.
(2)   It's raining. The streets are slippery.
(3)   Narrow roads are dangerous.
(4)   The poisonous apple poisoned Snow White.
(5)   The kid playing with the match burned the house.

They all indicate a causal relation. A causal relation usually consists of three elements of cause, effect and causal mark. For example, in (1), "john is poisoned" is the effect of "he ate a poisonous apple" and "because" is the causal mark. As we can see in the above examples, the causal elements may be words, phrases or predicates and also would occur in one sentence or two.

---
*corresponding author

Causal relations can be categorized from different points of view; e.g. according to the appearance of causal mark in the sentence. The causal mark may appear in the sentence or not and in the case of appearance it may be ambiguous or not. For example, in (3), "danger" is the effect of "narrow roads", but we do not have any explicit causal mark. Or in (4), the causal mark ("after") would be ambiguous as it is not always an indicator of causality.

On the other hand, the causative constructions may be either explicit or implicit (Girju, 2003). Even cause or effect may be expressed implicitly in a causal sentence. E.g. in (5) "kid playing with the match" is not the reason of burning the house, the fire is the reason. So finding the exact boundaries of causal relations is very important and challenging and needs world knowledge.

Recognition of causal elements can be done through rule-based methods or machine learning ones that need tagged corpus.

In general, it seems that corpus-based approaches are well suited for such problems (Goyal et al., 2017). Corpus-based causal detection systems are meant to determine the borders of each causal element within a single sentence or a sentence pair and classify them into predefined categories of cause, effect and causal mark. Causal training corpora already exist in languages such as English (some of them are introduced in section 2), but to the best of our knowledge such corpora are not developed for Persian. In this paper, we present a causality annotated corpus in Persian (PerCause) and a benchmark for training models based on the created dataset PerCause covers simple and general causation and complicated ones such as metaphysical causations, or weak, negated or nested causal relations are not in our scope.

This paper is organized as follows. Section 2 reviews related work, in section 3 we introduce Persian causality annotated corpus (PerCause) and explain the details of its production procedure. In section 4 we incorporate different machine learning and deep learning methods to extract causal relations based on PerCause and finally in section 5 the details of evaluations and discussion of the paper are presented. Section 6 concludes the paper.

## 2. . Related work

Here we review related researches in two main categories: 1) causality detection methods that may lead to semi-automatic development of a corpus and 2) manual creation of causality annotated corpora.

### 2.1 Causality detection

Causal relation extraction can be done through two approaches: rule based methods and machine learning methods. Some researchers use linguistic patterns to identify explicit causation relations. For example, Garcia (1997), employs a semantic model which classifies causative verbal patterns (as linguistic indicators such as <NP1 causative verb NP2>) to detect causal relationships in French texts. Khoo and colleagues (2000) use predefined verbal linguistic patterns to extract cause-effect information from business and medical newspaper texts. Lou and colleagues (2016) propos a framework that automatically harvests a network of causal-effect terms from a large web corpus using specified patterns (such as A leads to B, or If A Then B). Backed by this network, they propose a metric to model the causality strength between terms. In this method, causal relationships in short texts are identified using the created resource and also the presented statistical criteria.

Causality extraction by machine learning needs a causality corpus which is prepared manually or semi-automatically based on particular patterns. For example, in (Girju, 2003) a training corpus, containing sentences having 60 simple causal verbs is created using a LA TIMES section of the TREC 9 text. Using a syntactic parser, 6523 relations of the form NP1-Verb-NP2 are found, from which 2101 are manually tagged as causal relations and 4422 as non-causals. These make the positive and negative examples used to train a decision tree classifier to classify causal and non-causal relations in text.

Chang and Choi (2005) come up with a workaround mechanism to learn cue phrase and lexical pair probabilities from raw and domain-independent corpus in an unsupervised manner. They use these probabilities to extract inter-noun and inter-sentence causalities. In 2008, Blanco and colleagues (Blanco, et al., 2008) followed up with a work that considered explicit causal relations expressed by the pattern VerbPhrase-Relator-Cause, where relator $\in$ {because, since, after, as}. They used semantically annotated SemCor 2.1 corpus for training a C4.5 decision tree binary classifier with bagging.

Mirza and Tonelli in 2014 present CATENA, a system to perform temporal and causal relation extraction and classification from English text. To do this, they exploit the interaction between the temporal and the causal relations. They use a hybrid approach combining rule-based and supervised classifiers for identification of causal relations and adopt the notion of causality proposed in the annotation guidelines of the CausalTimeBank (Mirza et al., 2014; Mirza and Tonelli, 2014). This guideline accounts for CAUSE, ENABLE and PREVENT phenomena. Authors aim to assign a causal link to pairs of events when the causal relation is expressed by affect, link and causative verbs (CAUSE-, ENABLE- and PREVENT-type verbs) or when the causal relation marked by a causal mark

Mirza (2014) have proposed annotation guidelines for causality between events, based on the TimeML definition of events, which considers all types of actions (punctual and durative) and states as events. They introduced the <CLINK> tag to signify a causal link and also introduced the notion of causal signals through the <C-SIGNAL> tag.

Ning, et. Al (2018) claim that because a cause must occur earlier than its effect, temporal and causal relations are closely related and one relation often dictates the value of the other. So they present a joint inference framework for them, using constrained conditional models (CCMs), specifically, formulate the joint problem as an integer linear programming (ILP) problem, enforcing constraints that are inherent in the nature of time and causality.

In (Dasgupta, et, al, 2018) a linguistically informed recurrent neural network architecture (a bidirectional LSTM model enhanced with an additional linguistic layer) for automatic extraction of cause-effect relations from text is proposed. The architecture uses word level embeddings and other linguistic features to detect causal events and their effects mentioned within a sentence.

And finally in (Hashimoto, 2019) a weakly supervised method for extracting causality knowledge (such as Protectionism -> Trade war) from Wikipedia articles in multiple languages is presented. The key idea is to exploit the causality describing sections and multilingual redundancy of Wikipedia.

## 2.2 Manually annotated causality corpora

There are several manual annotated resources for causality. Verb resources such as VerbNet (Schuler, 2005) and PropBank (Palmer et al., 2005) include verbs of causation. Likewise, preposition schemes (e.g., Schneider et al., 2015, 2016) include some purpose- and explanation related senses. But these cases just cover specific classes of words.

Up to the best of our knowledge there is no causality annotated corpus for the Persian language. But there are few causality manual annotated datasets for English language such as: BECAUSE (The Bank of Effects and Causes Stated Explicitly) (Dunietz, et al., 2015), BioCause (Claudiu et al, 2013), CaTeRS (Mostafazadeh et al., 2016) and (Dunietz, Levin and Carbonell, 2015).

BECAUSE distinguishes three types of causation, each of which has slightly different semantics: Consequence, in which the cause naturally leads to the effect (e.g. We are in serious economic trouble because of inadequate regulation.); Motivation, in which some agent perceives the cause, and therefore consciously thinks, feels, or chooses something (e.g. we don't have much time, so let's move quickly); and Purpose, in which an agent chooses the effect because they desire to make the cause true (e.g. coach them in handling complaints, so that they can resolve problems).

CaTeRS focuses on nine causal relations combined with temporal relation, including: cause' (before /overlaps) 'enable' (before/overlaps), 'prevent' (before/overlaps), cause-to-end' (before/overlaps/during). In fact, CaTeRS incorporates four causal relations of cause, enable, prevent, and cause-to-end while assuming in 'A cause/enable/prevent B', that A's start is before B's start, but there is no restriction on their relative ending. This implies that a cause relation can have any of the two temporal relations: before and overlaps.

BioCause provides an annotation framework for causal relations in biomedical texts and marks the connective and argument spans and the direction of causality. (Dunietz, Levin and Carbonell, 2015) is a generalization of BioCause to broader domains and distinguishes four different types of causal relationships (Motivation, Inference, Purpose and Consequence).

Besides English corpora, there are similar corpora for other languages. Rehbein and Ruppenhofer (2020) present a new resource for German causal language. The annotation scheme is adapted from (Dunietz et al. 2015), but with an extended set of arguments. This dataset includes 4,390 annotated instances for more than 150 different triggers. The annotation scheme distinguishes three different types of causal events (Consequence, Motivation and Purpose). They also provide annotations for semantic roles, i.e. The cause and effect for the causal event as well as the actor and affected party. This research uses a BERT-based causal sequence tagger as the benchmark.

Also Salford Arabic Causal Bank (SACB) corpus (Sadek and Meziane, 2018) is a new corpus dedicated to Arabic Causal relations. It includes a collection of annotated sentences each of which contains an instance of a causal particle. This corpus is annotated with instances containing words prefixed with certain proclitic along with cause and effect arguments.

Statistics of these corpora are indicated in Table 1. The last row shows PerCause which is introduced in this paper to be compared to available corpora.

**Table 1.** *Statistics of causality corpora*

| Corpus | No of causal relations | Data Source Description | Relation types |
|---|---|---|---|
| BECAUSE | 1803 | 5380 sentences of NYTimes, Penn Treebank and other news articles | consequence, motivation and purpose types for causal relations |
| BioCause | 851 | 19 open-access full-text biomedical journals | Event trigger, type, theme and cause for each relation |
| CaTeRS | 488 | 320 stories (1,600 sentences) from the ROCStories corpus. | four types of causal relations of cause, enable, prevent, and cause-to-end while assuming in 'A cause/enable/prevent B' |
| Dunietz, Levin and Carbonell, (2015) | 400 | 1200 sentences from Washington section of the New York Times Corpus | Motivation, Inference, Purpose and Consequence types for causal relation |
| SemCore2.1[2] | 1068 | 352 texts from Brown Corpus | Causal/non-causal labels |
| German causal corpus | 4,390 | 150 text from two sources, (i) newspaper text from the TiGer corpus and (ii) political speeches from the Europarl corpus | three different types of causal events (consequence, motivation and purpose) |
| SACB | 2162 | 69573 tokens from arabiCorpus (Newspapers category containing approximately135 million words of articles published between 1996 and 2010 in different Arabic countries) | Causal/non-causal labels |
| PerCause | 5128 | 4446 sentences (129,000 tokens) from Bijankhan corpus and general books | Cause, Effect and Mark for each causal relation |

Also there are some causal domain specific corpora in biomedical domain which are dedicated to the diseases, symptoms and drugs such as CADEC (Karimi et al., 2015). This corpus consists of 1253 posts from AskaPatient Medical forum (7398 sentences) and determines drug, adverse effect, disease, symptom and finding in the text. There are other similar corpora (Leaman et al., 2009, Gurulingappa et al., 2010 and Deleger et al., 2012) in this field that due to their specific domain and tags are not so related to our work and we just mention them.

The other corpora related to causality are entailment datasets which include sentence pairs with entailment, contradiction or neutral relations. Causality is a form of entailment and so, some of the entailment relations in these datasets may be causality relation. For example, the sentences "John is not hungry anymore." and "John just ate lunch" have both entailment and causal relations.

---

[2] http://lit.csci.unt.edu/~rada/downloads/semcor/semcor2.1.tar.gz

The RTE[3] Competitions corpus include test and development categories from 2005 to 2010. These competitions have continued as TAC[4] and semEval competitions since 2010. SICK[5] corpus consists of 10,000 English sentence pairs, which are made up of two sources: ImageFlickr and SemEval2015 video descriptions, which are manually labeled with three labels of entailment, conflict, and Neutral.

And finally Stanford University SNIL[6] corpus consists of 500,000 pairs of the sentences, which is annotated manually in three categories of entailment, conflict, and Neutral.

There are some issues with the corpora discussed in most of the above researches. The first one is that the authors assume that a causal element is continuous and there is no gap between its words but sometimes a causal element (e.g. cause) may be split into two or more parts (words or tokens) which are not neighbors in the sentence. The example is presented in table 3. Secondly the considered patterns are limited to just causal verbs as causal marks or a number of limited grammatical patterns. These may lead to some problems for free word order languages like Persian. For example, "سیگار احتمال سرطان را بالا می برد/smoking increases the possibility of cancer" is a causal sample that simple causal patterns cannot cover.

In this paper, we discuss our methodology to prepare a causality corpus with two major properties: (1) there is no predefined limitation in grammatical patterns, this means that any causal pattern may occur in PerCause and (2) causal elements may have more than one part and the parts may have distance in the sentence so there is no limitation on their occurring location in the sentence. The corpus is IOB-tagged (Ramshaw and Marcus, 1995) for causal elements.

## 3. The causality annotated corpus

PerCause, our developed causality annotated corpus, consists of almost 129,000 tokens and 5128 causal relations. The initial dataset and tagset used in the corpus are introduced in the following subsections.

### 3.1 Data source

The initial corpus is selected among two different corpora: 1) Peykareh corpus and 2) Books corpus. Peykareh corpus (Bijankhan, 2005) consists of almost 10 million POS-tagged tokens (we just used the raw corpus). Although Peykareh corpus contains texts from different genres, news articles make the majority of it and as news articles often do not properly cover general causal elements, we gathered 10 Persian novels and general books and constructed a general corpus consisting 1.8 million tokens besides Peykareh corpus. The book corpus is not POS tagged and we used an automatic tagger (explained in section 4.2) when needed.

### 3.2 Tagset

As the concept of causality requires, the tagset contains 3 tags of "cause", "effect" and "causal mark". Due to the similarity of causality annotation to NER or chunking task, IOB format is selected for this task. Table 2 shows definitions and examples for the three tags. Some annotation examples are indicated in Table 3 and figure 1.

---

[3] Recognizing Textual entailment
[4] Text Analysis Conference
[5] Sentences Involving Compositional Knowledge, available in: http://clic.cimec.unitn.it/composes/sick.html
[6] http://nlp.stanford.edu/projects/snli/

**Table 2.** *Tags definitions and examples*

| Tag | Definition | Example |
|---|---|---|
| Cause | A person, event, action or thing that gives rise to an action, phenomenon, or condition. | Smoking increases the risk of cancer.<br>سیگار کشیدن ریسک سرطان را بالا می برد. |
| Effect | A change which is a result or consequence of an action or other cause. | Smoking increases the risk of cancer.<br>سیگار کشیدن ریسک سرطان را بالا می برد. |
| Causal Mark | A word or phrase which marks causal relation. | Smoking increases the risk of cancer.<br>سیگار کشیدن ریسک سرطان را بالا می برد. |

**Table 3.** *Labeled data sample*

کمبود هورمون کورتیزول باعث خشونت طلبی در برخی از پسران می شود، خصوصا در سنین نوجوانی.

<entity="cause-1">کمبود هورمون کورتیزول</entity> <entity="Causal-mark-1">باعث</entity> <entity="effect-1">خشونت طلبی در برخی از پسران</entity> می شود، <entity="effect-1">خصوصا در سنین نوجوانی</entity>.[7]

Cortisol hormone deficiency especially in teenage years, causes violence in some boys.

<entity"=cause-1">Cortisol hormone deficiency </entity> <entity"= effect-1"> especially in teenage years</entity>, <entity"=Causal-mark-1"> causes </entity> <entity"= effect-1">violence in some boys</entity>.

```
1  Cortisol hormone deficiency especially in
2  teenage years, causes violence in some boys.
3
4  Cortisol   → B-Cause
5  hormone    → I-Cause
6  deficiency → I-Cause
7  especially → B-Effect
8  in         → B-Effect
9  teenage    → I-Effect
10 years      → I-Effect
11 ,          → O
12 causes     → B-Mark
13 violence   → I-Effect
14 in         → I-Effect
15 some       → I-Effect
16 boys       → I-Effect
17 .          → O
```

```
۱  کمبود هورمون کورتیزول باعث خشونت طلبی در
۲  برخی از پسران می شود، خصوصا در سنین نوجوانی.
۳
۴  کمبود     → B-Cause
۵  هورمون    → I-Cause
۶  کورتیزول  → I-Cause
۷  باعث      → B-Mark
۸  خشونت طلبی → B-Effect
۹  در        → I-Effect
۱۰ برخی      → I-Effect
۱۱ از        → I-Effect
۱۲ پسران     → I-Effect
۱۳ می شود    → O
۱۴ ،         → O
۱۵ خصوصا     → I-Effect
۱۶ در        → I-Effect
۱۷ سنین      → I-Effect
۱۸ نوجوانی   → I-Effect
۱۹ .         → O
```

**Figure 1.** *IOB format sample*

---

[7] The Persian language is written from right to left

*3.3 Semi-automatic preparation of initial causality corpus*

After preparing the initial raw corpus from the combination of Peykareh and the book corpora, we choose a subset of corpus for manual tagging through an automatic method. In this section, the processes performed to construct this subset called initial causality corpus are expressed.

After applying these steps, the corpus is ready for IOB tagging and causal elements annotation.

**-Normalization**

As some characters in Persian keyboards have more than one corresponding Unicode (like 'ی' and 'ک'), in this step, characters are unified.

**-Sentence segmentation**

After normalization, text is segmented into sentences. We use punctuations such as ".", "!" and "?" as sentence delimiters while we concern using these symbols within special words (such as acronyms), numbers and URLs to avoid mis-segmentations.

**-Tokenization**

Tokenization converts each sentence into a sequence of tokens including words, punctuation marks, numbers, etc. As space is not a deterministic delimiter in Persian and it may occur within a word, or two separate words may appear without any space in between, tokenization is more challenging in this language comparing to English. In this stage, we use STeP-1 toolkit (Shamsfard et al., 2009) for tokenizing sentences.

**-Length filtering**

After segmenting the text into sentences and sentences into tokens, we remove very long and very short sentences and just keep sentences with length between 5 and 100 tokens. Our observations show that sentences with less than 5 or more than 100 tokens either does not contain any causal relation or are probably made by incorrectly merging or splitting of constituents and therefore can be ignored. Sentence loss rate is discussed at the end of this section.

**-Causal candidate selection**

Here, sentences that conform to predefined patterns of causality are called causal candidates. We select final set for manual labelling through this candidate set. In this regard, we developa list of 30 causal patterns (or somehow cue phrases), some of which are indicated in table 4. In table 4, # means that the pattern words should be seen in the defined order but, they may not necessarily make a phrase.

At the **first point**, for making this list, we manually **prepare** two lists of 1) initial patterns (about 10 causal patterns) and 2) initial causal pairs (about 80 pairs of causes and effects). Then, we search initial causal pairs through the preprocessed corpora and extract sentences containing these causal pairs. On the other hand we search initial patterns in the corpus and extract more causal pairs from the extracted sentences. Thenrepeat the first step again with new extracted causal pairs and continue the iteration. The list of causal patterns came from investigating the structure of sentences containing causal patterns one by one manually. The set of sentences that match the causal patterns are selected as initial causal candidates. This procedure is presented in Figure 2.

> 1- Make initial pattern list and initial causal pairs list manually
> 2- While any new causal pair or causal pattern is found
>     a. search causal pairs list through the corpus and extract more causal patterns
>     b. search causal patterns list in the corpus and find new causal pairs
>     c. update both lists
>
> 3- Select set of sentences that match the causal patterns as initial causal candidates for manual labelling.

**Figure 2.** *Pseudocode of causal candidate sentence selection for labelling*

**-Final sentence set selection for tagging**
After preparing the initial causal candidate set, the extracted pairs are revised and inappropriate ones are removed . In fact, some cause-effect pairs are not a permanent pair and just occurr to be causal in a specific sentencee.g. "He went to Paris because he was ill".This kind of sentences in news articles are seen frequently. As we want the sentences to be general and have high accuracy in all contexts, we mainly are looking for examples that are direct cause of the effect in the sentence, so we inspect samples in the corpus and prune some patterns and select a final set for tagging.
There is no sentence loss during the first 3 steps of preprocessing (Normalization, tokenization and sentence segmentation). But in the length filtering step we have about 10 percent loss. After causal candidate selection (6% of initial sentences have causal patterns), we select about 50 percent of sentences as causal candidates having causal relations that always are hold.
Table 5 shows the specification of the initial corpus before causal labelling.

**Table 4.** *Causal patterns examples*

| Pattern category | Patterns | Percentage |
|---|---|---|
| Verbs | change, increase, lead to, raise, cause, produce<br>(تولید کردن، افزایش دادن، منجر شدن، بالا بردن، موجب#شدن، تغییر دادن) | 23% |
| Prepositions | Through, if#then, because, as a result, since<br>(به وسیله، اگر#آنگاه، چون ، درنتیجه، زیرا، از آنحایی که) | 5% |
| Nouns | Result, reason, cause<br>(حاصل، سبب، باعث، موجب) | 45% |
| Hybrid | reason of %, because of, caused by, result of this, reason#percent, using#for, for this reason, high#probability, important#reasons, because of this<br>(علت#٪ ، به این بابت، به علت، به دلیل، حاصل از، حاصل#این#آن، علت#درصد ، از#برای#استفاده ، ناشی از، به همین علت ، به همین دلیل، احتمال#افزایش ، عوامل#مهم، به این دلیل) | 27% |

**Table 5**. Specification of the corpus before manual causal labeling

|  | # of sentences | # of tokens |
|---|---|---|
| initial causal candidate corpus | 4446 | 129293 |

### 3.4 Tagging the causality corpus

After these steps, Sentences in this set are ready for being tagged

For this task we developed an annotation tool to help linguists annotate the sentences manually. This tool allows the annotator to tag a multi-token element (even with distance in between) and can be used by multiple annotators simultaneously. A page of its GUI is shown in figure 3. Annotators can just right click on a phrase or word and select the corresponding tag. Also annotator can tag a multi part element using entity number option. After Annotation, data is converted to IOB format in internal structure for further use.

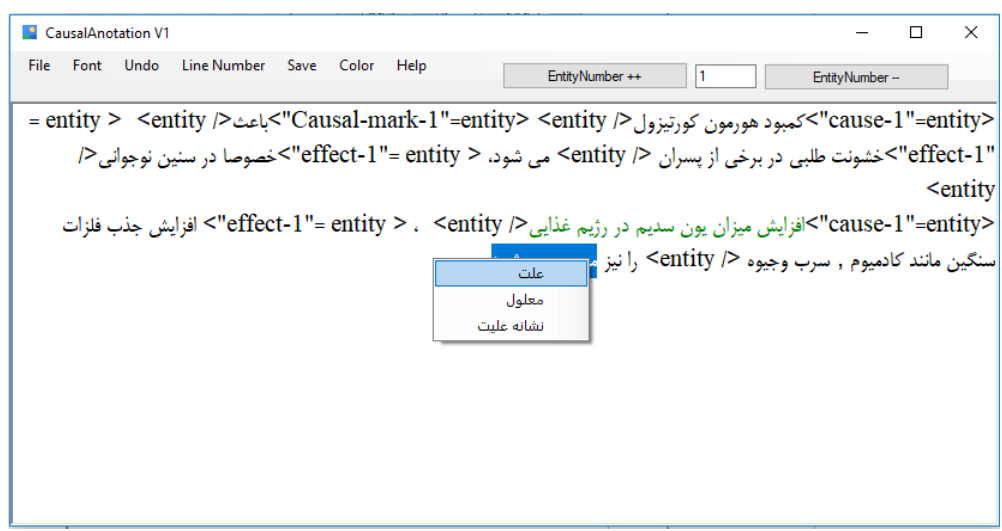

**Figure 3.** *GUI of annotation tool*

. Table 6 shows statistics of causal tags (B-Cause, I-Cause, B-Effect, I-Effect, B-Mark and I-Mark) in the corpus. We have about 63,000 causal tokens (tokens having causal tags) from 133293 total tokens of the corpus in the form of 5128 causal relations (causal triples or tuples) in 4446 sentences. Some candidate sentences may contain no causal relation.

### 3.5 Inter annotation agreement

An important factor in developing a corpus, is the inter-annotator agreement. As discrepancy between different annotators is inevitable in manual annotation of a corpus and it is even observed in practice that an annotator would tag a single text differently at two different times, we need to calculate inter annotation agreement to be sure about the accuracy of annotation results. On the other hand, experiences have shown that use of more uniform and precise corpus has led to better results in machine learning based systems. In this regard, different criteria are used to measure the degree of annotator agreement. One of the famous and widely used metrics for this issue is Kappa (Cohen, 1960).

**Table 6**. *Tag distribution in the corpus*

|   | tag | Number of occurrences | percentage |
|---|---|---|---|
| 1 | B-cause | 4261 | |
|   | I-cause | 19683 | |
|   | **Total cause** | **23944** | **%17.96** |
| 2 | B-Effect | 4597 | |
|   | I-Effect | 27180 | |
|   | **Total effect** | **31777** | **%23.84** |
| 3 | B-Causal mark | 5128 | |
|   | I-Causal mark | 2157 | |
|   | **Total causal mark** | **7285** | **%5.4** |
| 4 | Total tagged | 63006 | %47.27 |

This criterion is a statistical criterion used in corpus linguistics to examine the degree of agreement between two annotators of a corpus. In this metric, attempts have been made to reduce the effect of labels which are appear to be accidentally similar to each other. For this purpose, kappa is calculated as follows.

$$k = \frac{\Pr(a) - \Pr(e)}{1 - \Pr(e)}$$

In this formula, Pr(a) is the correct level of similarity between two annotators, and Pr(e) is the rate of accidental similarity between the two annotator. To use the Kappa criterion in cases with more than two annotators, this value is calculated for each pair of annotators and their mean values are taken.

In the field of linguistics, the kappa coefficients above 80% are considered to be appropriate for a corpus (Green, 1997).

Here, for reducing ambiguities in labelling procedure, we have prepared a precise annotation guide and employed two annotators for the process. 1300 tokens of data are randomly selected and has been annotated twice. The inter-annotator agreement is calculated as 94.5% so the created annotated corpus is admissible and is now ready to be used by causality detection systems.

## 4. Causal element boundary detection system

In this research, the goal is to construct a causal corpus and to make a benchmark for detecting causal pairs and their boundaries using the constructed corpus. For this purpose, we use the created causal corpus to train machine learning baselines and deep learning systems and make a benchmark in this task for research community. In this section we discuss the classification methods and features we employ to build the benchmark.

In fact, we address the causal element boundary detection problem as a multi-class classification problem that its classes are causal elements tags. Hence this section explains the classification part and the features which are fed to the system.

### 4.1 Classification methods

To develop the benchmark, a training dataset is separated and some classifiers such as Naïve Bayes, RBF (Radial basis function) and CRF are tested. Also we implemented two deep learning systems whit different representations; which are trained and tested using the same datasets.

1) A deep learning framework[8] (bi-LSTM+CRF in Tensorflow based on (Huang et. al, 2015). As is indicated in figure 4, the procedure is as followed:

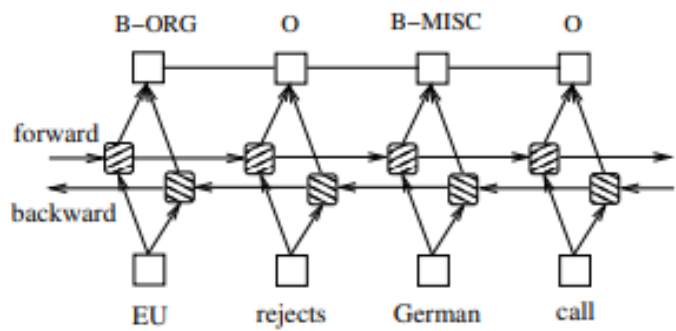

**Figure 4.** BI-LSTM-CRF model (Huang et, al, 2015)

- Concatenate final states of a bi-LSTM on character embeddings to get a character-based representation of each word

- Concatenate this representation to a standard word vector representation

- Run a bi-LSTM on each sentence to extract contextual representation of each word

- Decode with a linear chain CRF.

In this step, for word embedding, fast-text CBOW method on Persian Wikipedia corpus with more than 1 billion tokens is used.

2) Due to the emergence of Bert[9] models (Delvin, et al., 2018) and their positive performance in similar tasks, we fine-tune a pre-trained model on produced data as well.

Our results and implementation details on general classifiers and deep systems are presented in section 5.2.

### *4.2* **Features**

To train causal element boundary recognition model, major features of word, stem and part-of-speech tags have been used. We use stem feature mainly for covering various verb inflections in sentences. Also sometimes placement of specific POS tags adjacent to causal mark indicate a causal relation pattern. So we use POS feature too. These features are fed into classifiers as feature functions in the train stage.

---
[8] https://github.com/guillaumegenthial/sequence_tagging
[9] BERT: Pre-training of Deep Bidirectional Transformers for Language

We use two POS tagging models of 16-tags and 100-tags that were trained over Peykareh corpus with precision of 94% and 90% respectively. Also for stemming part, we use a light affix removal stemming algorithm which simply removes prefixes and suffixes. The effect of considering POS tags and stemmed words are evaluated in section 5.3.

## 5. Experimental results

### 5.1 Data and metrics

For covering the effect of size of data in our analysis, we performed our experiments on two data sets considering available labelled data: 1) first data pack with 47000 tokens and 2) second data pack with 129000 tokens (including the first pack)

For the first pack, we divide our tagged dataset into 3 parts of train, test and development with the portions of 80, 10 and 10 percent respectively. But for the second pack from non-overlapping part, we just separate a development set and add the rest to the train dataset. We used the same test dataset which contains 4750 tokens. Also we used 10-fold cross validation procedure to ensure robustness of the results.

Standard metrics of accuracy (Acc), precision (P), recall (R) and F-measure (F) are used to evaluate system performance which are defined as bellow. We calculate these metrics for each tag and consider the mean value of all 7 classes (B-Cause, I-Couse, B-Effect, I-effect, B-Mark, I-mark and O) for total comparison of systems.

$$\text{Accuracy(Acc)} = \frac{\text{all correct system decisions for specific tag}}{\text{all samples}}$$

$$\text{Precision(P)} = \frac{\text{correct system decisions for specific tag}}{\text{all system decisions}}$$

$$\text{Recall(R)} = \frac{\text{correct system decisions for specific tag}}{\text{what system should have decided}}$$

$$F - measure = \frac{2 \cdot P \cdot R}{(P + R)}$$

### 5.2 Evaluation of different machine learning methods on the Causality corpora (PerCause)

For General classifiers (RBF and Naïve Bayes) we used WEKA[10] toolkit with their default parameters and we converted tokens to word vectors.
For CRF classifier, we used PocketCRF and set f (frequency threshold) to 4 according to development procedure and set m=1 and p=4 for performance improvement (m=1 uses less memory and p determines thread number). Also we tested different window sizes.

---

[10] https://www.cs.waikato.ac.nz/ml/weka/

In first deep learning implementation (bi-LSTM+CRF), we used 45 epochs, 100 dimensions for character embedding and 300 dimensions for word vectors.

For second deep system (Bert-base), we used ParsBert (Farahani, et al., 2020) and fine-tuned the model with our causal data. For fine-tuning we used (weizhepei, 2020) implementation which is a PyTorch solution of Named Entity Recognition task with Google AI's BERT model. In weizhepei research, the F-measure result in NER task stated as 94.62 on Chinese MSRA and 96.4 on English Conll2003 dataset.

Tables 7 and 8 show the performance evaluation of different machine learning methods trained on the first and second data packs. Table 9 indicates best system results associated to different train data sets.

**Table 7.** *Comparison of ML systems performance for the first data pack (**47000 tokens**)*

| System | Precision | Recall | F-measure | Acc |
|---|---|---|---|---|
| CRF | **0.77** | **0.75** | **0.755** | 0.747 |
| RBF Network | 0.464 | 0.50 | 0.48 | 0.499 |
| Naïve Bayes | 0.466 | 0.48 | 0.472 | 0.479 |
| Deep Learning (same train set) | 0.65 | 0.52 | 0.58 | **0.86** |
| DeepLearning-ParsBert | ۰٫٦٤ | ۰٫٧١ | 0.6۷ | ۰٫٨٨ |

**Table 8.** *Comparison of ML systems performance for second data pack (**129000 tokens**)*

| System | Precision | Recall | F-measure | Acc |
|---|---|---|---|---|
| CRF | **0.799** | **0.72** | **0.761** | 0.755 |
| RBF Network | 0.493 | 0.479 | 0.462 | 0.478 |
| Naïve Bayes | 0.517 | 0.506 | 0.503 | 0.506 |
| Deep Learning (same train set) | 0.79 | 0.65 | 0.713 | **0.914** |
| DeepLearning-ParsBert | ۰٫٦6 | ۰٫٧7 | 0.706 | ۰٫٨٩ |

As table 7 and 8 indicate, if we consider total performance, the CRF system outperforms the other implemented systems with F-measure of 0.76 which is comparable to Girju (2003)'s 0.8, Chang and Choi (2005)'s 0.81 and Arabic ML system of (Sadek and Meziane, 2018) with best result of 0.76 on their own data. Details of CRF evaluation is presented in the next section.

For deep systems, (Rehbein and Ruppenhofer, 2020) reached the average of 72.2 F-measure on determining causal labels in their German corpus which is comparable to our deep learning systems. Although in (weizhepei, 2020) Bert-based implementation on similar task (NER), the F-measure result stated as 94.62 on Chinese MSRA and 96.4 on English Conll2003 dataset which these results have definitely happened due to the large and rich corpus and possibly because of the absence of distant dependencies and non-breaking of entities into several parts.

The main point which is indicated in the above tables is that increasing data size does not have any observable change in CRF results, but it improves deep learning results by 13% which is remarkable.

In Bert based system, the final results are very close to other deep learning system, however the system performance with first pack and second data packs are nearly similar. This result confirms that Bert is functioning well with low data or the amount of data increase in the second pack has not been enough to improve the Bert system performance. In addition, it seems that the presence of the CRF layer in other deep-trained networks has made it slightly superior in F-measure, at least in this volume of training data.

*5.3 Studying the effect of automatically extending the corpus*

In the previous section we saw that although deep learning has the best accuracy among other systems, its performance (according to F-measure) is less than CRF. As deep learning methods basically need large scale data, we decided to produce data automatically using our most accurate system (bi-LSTM+CRF architecture) to study the effect of automatically enlarging the corpus size. For this purpose, bi-LSTM+CRF is selected to generate the automatic data because accuracy is more important than recall here, so we labeled additional data automatically with deep learning system.

Then we used the output generated by the bi-LSTM+CRF system to create a corpus of 400,000 tokens with accuracy of 0.91 (called automatic data pack). The new automatically generated data set is itself used to train a deep learning system. Data used for automatic annotation, is selected among Peykareh and news corpora and all sentences are from the same domain as PerCause.

Table 9 shows the results obtained on different data sets. In this table, in the last three rows, we used the automatically generated corpus with accuracy of 91%.

**Table 9.** *Comparison of deep learning performance for different train data sets*

| System | F-measure | Acc |
|---|---|---|
| First data pack | 0.587 | 86 |
| Second data pack | **0.713** | **91.4** |
| Automatic created data pack | 0.561 | 84.47 |
| First pack + automatic data pack | 0.576 | 80.37 |
| second pack + automatic data pack | 0.67 | 89.7 |

As Table 9 indicates, adding automatically generated data did not improve system performance. This is probably because this produced data pack is not precise enough to use as training data. In future work to use this automatic database, we will consider weakly supervised methods.

*5.3 Best system evaluation*

As in the previous section has been showed, with current size of data, among all tested systems, CRF has the best total performance, although deep architecture has better accuracy. This result is mostly expected, because the CRF has shown the best F-measure as compared to other machine learning methods in such problems.

In this section we study the effect of various features on the performance of CRF causality detection baseline, for each class. In the table 9 the results of CRF system without any additional features (just words and labels) and with two simple features is presented for each class.

**Table 10.** *CRF system performance for each label according to features and window size*

| Features | Just words | | | Words+POS+stem | | | Words+POS+stem | | |
|---|---|---|---|---|---|---|---|---|---|
| Window size | 5 token | | | 5 token | | | 3 token | | |
| Label/Metric | P | R | F | P | R | F | P | R | F |
| B-cause | 0.755 | 0.57 | 0.65 | 0.72 | 0.65 | 0.68 | 0.75 | 0.69 | 0.68 |
| I-cause | 0.598 | 0.59 | 0.59 | 0.61 | 0.71 | 0.66 | 0.62 | 0.65 | 0.63 |
| B-effect | 0.80 | 0.68 | 0.74 | 0.82 | 0.72 | 0.77 | 0.82 | 0.69 | 0.75 |
| I-effect | 0.79 | 0.73 | 0.76 | 0.84 | 0.74 | 0.79 | 0.79 | 0.71 | 0.75 |
| B-mark | 0.93 | 0.84 | 0.88 | 0.96 | 0.92 | 0.94 | 0.92 | 0.86 | 0.89 |
| I-mark | 0.82 | 0.69 | 0.75 | 0.71 | 0.77 | 0.74 | 0.66 | 0.77 | 0.714 |
| O | 0.69 | 0.76 | 0.73 | 0.71 | 0.75 | 0.73 | 0.72 | 0.78 | 0.75 |
| Average | 0.77 | 0.69 | 0.73 | **0.77** | **0.75** | **0.76** | 0.75 | 0.73 | 0.74 |

It seems that changing the size of adjacency window in CRF classification procedure, can directly affect the performance. Our best result is obtained through 5 tokens window. We have repeated our experiments with and without POS and stem features.

As table10 shows, adding POS and stem (Adding the stem feature slightly - by 0.5 units. - improves the total results.) Features improves system performance for 3 percent and our best performance is obtained for 5 tokens window.

For better indication, in figure 5 and 6, two charts are presented so we can clearly compare results of CRF system according to window size and features.

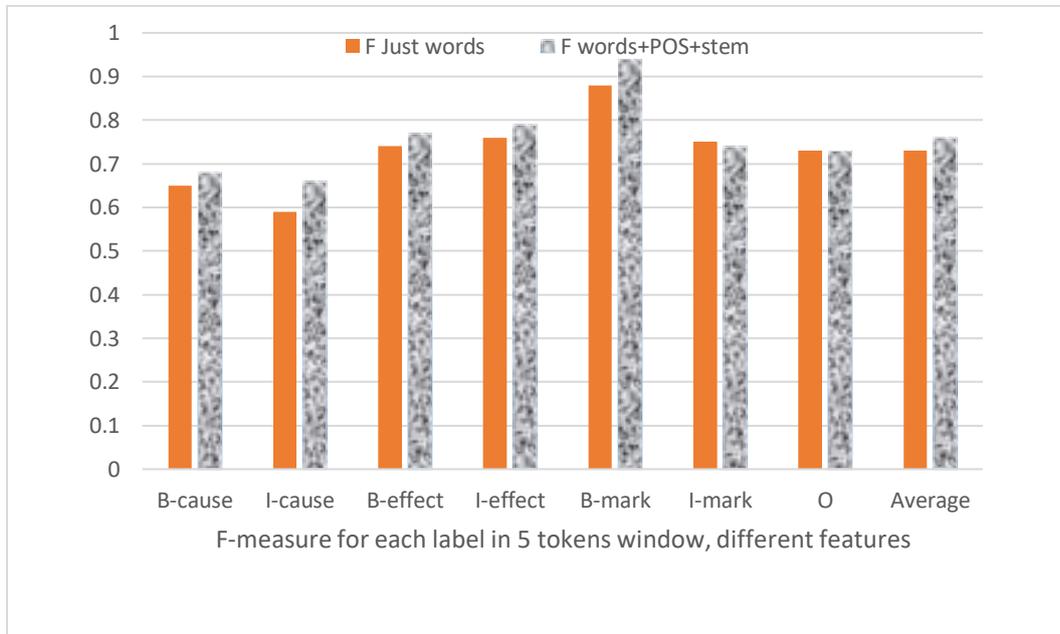

**Figure 5.** *CRF system performance; F-measure for each label in 5 tokens window and different features*

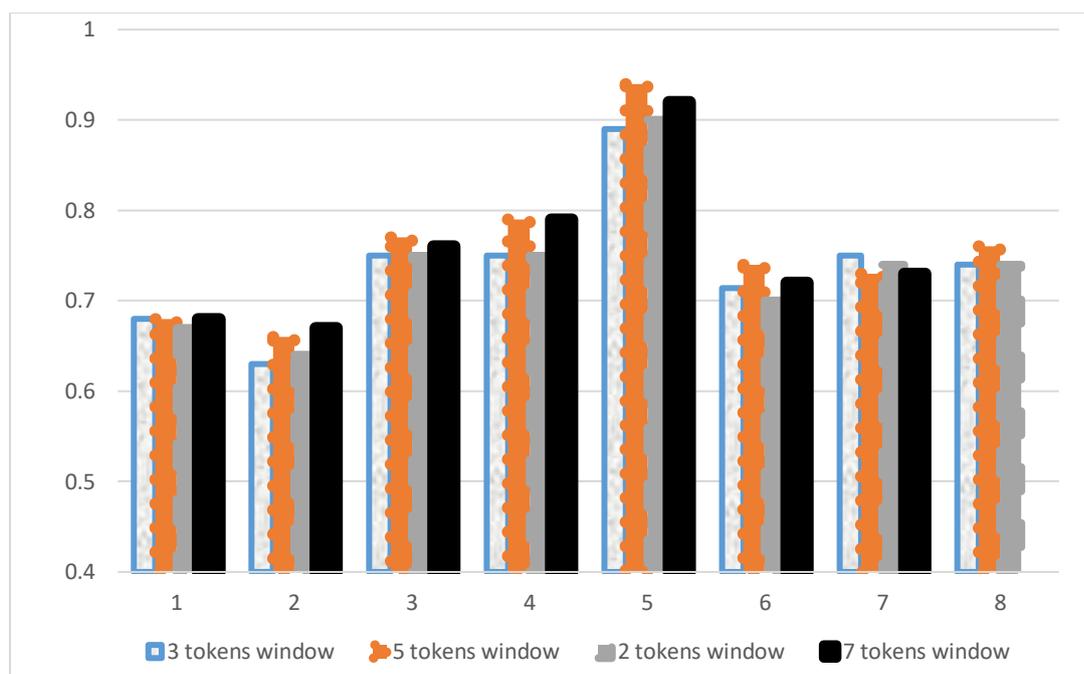

**Figure 6.** *CRF system performance; F-measure for each label in different size of window*

### 5.3 Discussion

As we observe above, CRF method has the best total results (F-measure) among the other baselines. The CRF algorithm is based on the conditional probability and graph theory. In contrast to many other methods, the conditional random technique considers each sample in the context of the tags of previous words, rather than the sample alone. And it may be the strength of this method among others. Also in CRF system, experimental results indicate that window size and basic features (POS and stem) are two effective parameters that can improve system performance.

On the other hand, we expected that deep learning method would have the best total performance, but probably due to the insufficient dataset the algorithm didn't outperform the others, although it has the best accuracy of 91.4%. Experiments show that size of train data has an obvious effect on the results in deep learning system, because after increasing data size (nearly we added twice of initial size to the first pack) the deep system F-measure has increased by 13 percent, meanwhile it did not affect CRF performance so much. To sum up, it seems that with small size of data it is better to use CRF classifier, but in the large scale data size deep learning would be a better choice.

Also the automatically created train corpus, apparently mislead the deep network and didn't lead to better results. Moving towards weak supervision may handle this issue.

Since our initial corpus size is small for a learning procedure, we use a coarse grain data set in which all types are causality are integrated in a general label of "causal relation". With bigger dataset we may have fine grain labels such as purpose and motivation in the future.

## 6. Conclusion

In this paper the production process of a Persian causal annotated corpus is presented. Furthermore, three machine learning methods are used for training a model for determining causal elements boundaries using this prepared data set. Among tested algorithms, CRF method with F-measure of 0.76 has the best total performance (according to F-measure), which is probably through its property of modeling context and not only the current sample. Meanwhile we also tested two deep learning architecture (a bi-LSTM+CRF and a Bert-base) which the first one has the best accuracy among other tested systems, but probably due to the insufficient size of data the total result (f-measure) is not as expected and unfortunately automatic generated data did not perform well too.

As future work, adding post processing and manual modification of automatic-generated dataset or moving toward weakly supervised methods may improve the results. Also adding other features may improve CRF system.